\documentclass[letterpaper]{article} 
\usepackage{aaai23}  
\usepackage{times}  
\usepackage{helvet}  
\usepackage{courier}  
\usepackage[hyphens]{url}  
\usepackage{diagbox}
\usepackage{graphicx} 
\usepackage{natbib}  
\usepackage{lipsum} 
\usepackage{caption} 
\usepackage{xcolor}
\usepackage{algorithm}
\usepackage{algorithmic}
\usepackage{balance}

\usepackage{graphicx}
\usepackage{amsmath}
\usepackage{amssymb}

\usepackage{newfloat}
\usepackage{listings}
\DeclareCaptionStyle{ruled}{labelfont=normalfont,labelsep=colon,strut=off} 
\floatstyle{ruled}
\newfloat{listing}{tb}{lst}{}
\floatname{listing}{Listing}
\title{Rethinking interpretation:\\ Input-agnostic saliency mapping of deep visual classifiers}
\author{
    Naveed Akhtar\textsuperscript{\rm 1},
     Mohammad A. A. K. Jalwana\textsuperscript{\rm 1}
}

\affiliations{
    \textsuperscript{\rm 1} The University of Western Australia, 35 Stirling Highway Crawley 6009, Australia.\\ naveed.akhtar@uwa.edu.au, asimjalwana@gmail.com

}
\usepackage{bibentry}

\begin{document}

\maketitle

\begin{abstract}
Saliency methods provide post-hoc model interpretation by attributing input features to the model outputs. Current methods mainly achieve this using a single input sample, thereby failing to answer input-independent inquiries about the model. We also show that input-specific saliency mapping is intrinsically susceptible to misleading feature attribution. Current attempts to use `general' input features for model interpretation assume  access to a dataset containing  those features, which biases the interpretation.  Addressing the gap, we introduce a new perspective of input-agnostic saliency mapping that computationally estimates the high-level  features attributed by the model to its outputs. These features are geometrically correlated, and are computed by accumulating model's gradient information with respect to an unrestricted data distribution. To compute these features, we nudge independent data points over the model loss surface towards the local minima associated by a human-understandable concept, e.g.,~class label for classifiers. With a systematic projection, scaling and refinement process, this information is transformed into an interpretable visualization without compromising its model-fidelity. The visualization serves as a stand-alone qualitative interpretation. With an extensive evaluation, we not only demonstrate successful visualizations for a variety of concepts for large-scale models, but also showcase an interesting utility of this new form of saliency mapping by identifying backdoor signatures in compromised classifiers.
\end{abstract}

\section{Introduction}
Deep perceptual models are at the heart of many recent scientific developments~\cite{lecun2015deep}, \cite{Nat_online}. Their applications now reach beyond the rudimentary decision making, to the automation of high-stake domains, e.g.,~self-driving vehicles~\cite{deng2021deep}, smart security~\cite{DARPA_AINext}, health-care~\cite{tang2021edl}. This is a direct consequence of their established  human-level  ability to discriminate between intricate visual patterns~\cite{lecun2015deep}. 
However, like any deep learning approach, they are black-box techniques~\cite{vinuesa2021interpretable}. It is normally very hard to interpret what information has been learned by these models and how it influences their predictions.

\begin{figure}[t]
    \centering
    \includegraphics[width = 0.29\textwidth]{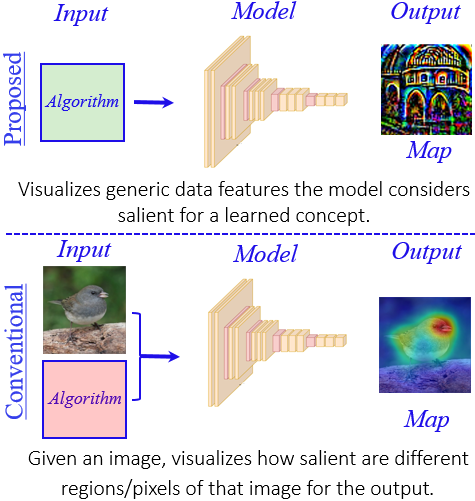}
    \caption{Difference between the proposed and conventional saliency mapping. (\textbf{Top}) We propose to compute maps that are input-agnostic and can attribute outputs to generic data concepts. (\textbf{Bottom}) Conventional approaches allow interpretations that are  specific to a given input. }
    \label{fig:teaser}
\end{figure}

Owing to the significance of the perceptual models, their interpretability is currently  considered a mainstream problem in computer vision~\cite{fong2019understanding}, \cite{fong2017interpretable}, \cite{petsiuk2018rise}, \cite{zeiler2014visualizing}, \cite{jalwana2021cameras}, \cite{selvaraju2017grad}. In the literature, it has been translated to the task of identifying the input features deemed salient by the model to make its predictions. 
Mainly two broad strategies are popular to accomplish this task. The first searches for salient features in an input image by selectively perturbing its different regions and analyzing the resulting effects on the model predictions~\cite{fong2019understanding}, \cite{fong2017interpretable}, \cite{petsiuk2018rise}, \cite{zeiler2014visualizing}. Though effective, these methods may require heuristics to control their search space, potentially compromising the model-fidelity of the map due to the external influence~\cite{jalwana2021cameras}. 

\begin{figure*}[t]
    \centering
    \includegraphics[width = 0.77\textwidth]{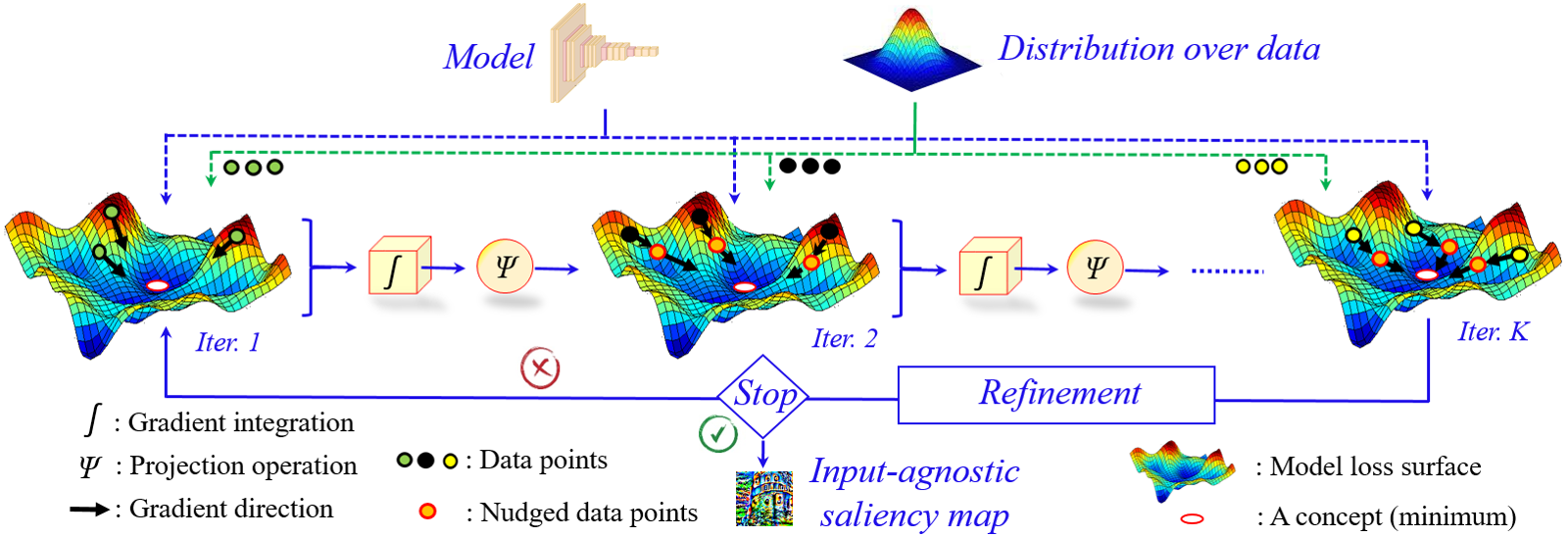}
    \caption{\textit{Central idea}: We iteratively integrate and refine the gradient information of the model's loss surface to estimate an input-agnostic map that captures geometric input features considered salient by the model. An iteration draws \textit{i.i.d.}~samples from a distribution and nudges them to the nearby local minima that belong to a human-understandable concept (e.g.,~a class label). By integrating ($\int$) the gradient information from these nudges, and projecting ($\Psi$) it onto a norm-bounded surface, we amplify the salient geometric patterns w.r.t.~the model. A refinement is further employed to improve the visualization without compromising the model-fidelity of the computed map.}
    \label{fig:main}
\end{figure*}

The second strategy is commonly known as backpropagation saliency mapping~\cite{selvaraju2017grad}, \cite{rebuffi2020there},  \cite{simonyan2014deep}, \cite{springenberg2014striving}, \cite{zeiler2014visualizing}, \cite{zhang2018top}. It estimates feature saliency by analysing the gradients and activations of the internal layers of the model for a given  input.  Low resolution map estimates, and failing the basic sanity checks, are known challenges faced by  this strategy~\cite{adebayo2018sanity}, \cite{jalwana2021cameras}. Leaving aside the peculiar issues of both  strategies, the eventual saliency map computed by the both are always \textit{input-specific}. Implying, the resulting interpretation is only valid for a single sample, see Fig.~\ref{fig:teaser}. This is too restrictive, especially when we consider that the ultimate objective of this research direction is  \textit{model} interpretation.   

In this work, we approach   saliency mapping from a different   perspective. Our objective is to map generic (as opposed to input-specific) features of the modelled data which are deemed salient by the model. Under this input-agnostic  perspective, the target map  is a human-understandable visualization of the salient features  attributed by the model to its outputs. The central idea of our technique is illustrated in Fig.~\ref{fig:main}. To compute the map, we iteratively explore the loss surface of the model by nudging a set of independent data points in the directions of nearby local minima  associated with the output. The  direction estimated by the nudges are integrated to characterize the landscape of the model's loss surface. By a gradual accumulation of this information and its projection onto a norm-restricted surface, we amplify the geometric correlation in the map which is subsequently  refined computationally.  When expanded to an image-grid, this map visualizes salient geometric features associated by the model to its output, i.e.,~a concept. 

A unique property of our saliency map is that it enables a holistic visualization of the model's understanding of its own  outputs. This can lead to many interesting applications. To illustrate one, this work also presents a case study to detect Trojan trigger patterns in compromised classifiers that contain backdoors~\cite{wang2022survey}. We leverage our input-agnostic visualization to map the geometric patterns to which the classifier's output nodes are more sensitive. A model that has a backdoor, leads to  visualisations that contain traces of the patterns used to trigger the backdoor, thereby allowing Trojan  detection. The contributions of this paper are as follows.
\begin{itemize}
    \item Systematically highlighting the limitations of input-specific saliency estimation, it introduces a new perspective of mapping generic data features attributed by a model to its outputs - input-agnostic saliency mapping. 
    \item It devises a first-of-its-kind method for the proposed input-agnostic saliency mapping using  model gradients. 
    \item It showcases a utility of the new saliency mapping with backdoor trigger identification in  classifiers.
    \item It introduces a quantitative metric for the newly proposed mapping framework and performs evaluation on large-scale ImageNet models to support the claims. 
\end{itemize}

\section{Related work}
For  perceptual model interpretation through saliency mapping, there are two popular streams of methods. The first  perturbs different regions of the input and records the effects of these perturbations to estimate the contribution of  regions to the model  prediction~\cite{fong2017interpretable}, \cite{fong2019understanding}, \cite{petsiuk2018rise}, \cite{sundararajan2017axiomatic}, \cite{erion2021improving}. The second uses activations of the deeper layers of the underlying neural network and  back-propagated model gradients to estimate a saliency map~\cite{selvaraju2017grad}, \cite{jalwana2021cameras}, \cite{simonyan2014deep}. Due to the rising importance of model interpretation in numerous applications, there is a wide interest of the community in developing methods along both the directions. Influential contributions for both are discussed below.

Among the \textit{perturbation-based saliency} methods, RISE~\cite{petsiuk2018rise} and Occlusion~\cite{zeiler2014visualizing} estimate the maps by weighting the perturbation masks with respect to the changes in the model confidence score. 
Other methods, for instance,  Extremal perturbations~\cite{fong2019understanding}, Meaningful perturbations~\cite{fong2017interpretable}, Real-time saliency \cite{dabkowski2017real} and \cite{ribeiro2016should},  cast the underlying problem into an optimization objective. Although the perturbation methods are generally effective, they do not specifically  shield the computed maps from external influence. A major  challenge behind this problem is that the search nature of these methods requires exploring a vast  solution space. Hence, the techniques rely on heuristics, priors or external constraints for tractability. This can affect the model-fidelity of the resulting interpretations~\cite{jalwana2021cameras}. 

Within the perturbation-based methods, there is also a branch of techniques that takes axiomatic approach towards saliency map creation~\cite{sundararajan2017axiomatic}, \cite{erion2021improving}, \cite{pan2021explaining}, \cite{srinivas2019full}. Defining a path from a baseline image to the input, this paradigm integrates the effects of perturbation to images along this path to compute a saliency map. Although techniques vary in defining the paths and signal integration schemes, all of them compute saliency map for a single image. These methods do provide certain desirable theoretical properties, however they also entail high computational cost for image-specific interpretations. 

The \textit{back-propagation saliency methods}~\cite{simonyan2014deep}, \cite{selvaraju2017grad}, \cite{zeiler2014visualizing}, \cite{jalwana2021cameras} are highly popular as perceptual model interpretation tools. Normally, they construct a saliency map for the regions of an input image. 
These methods are also relatively  computationally  efficient \cite{zintgraf2017visualizing}, \cite{kapishnikov2019xrai}.  \citet{simonyan2014deep} were among the first to use model gradients for interpretation. Numerous improvements to this idea have been subsequently proposed to handle the noise sensitivity of the model gradients. \citet{springenberg2014striving} proposed Guided back-prop, while \citet{zeiler2014visualizing}  altered the back-propagation rules for the ReLU layers of the model for that purpose.
Similarly, SmoothGrad~\cite{smilkov2017smoothgrad} computes the average gradients over the samples in the close vicinity of the original input  to mitigate the gradient noise  sensitivity. 

Along a similar line of thought, DeepLIFT \cite{shrikumar2017learning},  Excitation Backprop \cite{zhang2018top} and LRP \cite{bach2015pixel}, recast the back-propagation rules for saliency mapping to restrict the sum of attribution signal to unity. In another effort to control the signal noise, \citet{sundararajan2017axiomatic} combined multiple attribution maps. There are a number of methods that estimate the saliency map by merging layer activations of a model and its gradient information. Popular examples include CAM \cite{zhou2016learning}, linear approximation \cite{kindermans2016investigating}, GradCAM \cite{selvaraju2017grad}, NormGrad \cite{rebuffi2020there} and CAMERAS~\cite{jalwana2021cameras}. Though appearing late in the literature, these methods are currently dominating the backpropagation-based saliency mapping corpus of the literature. 

Both of the above streams have an obvious limitation. They both explain the model behavior using only the features present in the given input sample. Realizing this restriction, works such as \cite{bau2017network} and \cite{ghorbani2019towards} use `general' object features to explain the model. However, these general features are extracted from a dataset. Thus, the eventual explanation is intrinsically biased to that data. Moreover, similar to the above-discussed  streams, these methods must still use individual inputs to visualize the explanations. In this work, we devise a saliency mapping technique that truly liberates model explanation from the input, resulting in a first-of-its-kind input-agnostic saliency mapping mechanism.


\section{Proposed saliency mapping}
To  motivate the idea and relate it to the existing practice, we start our  discussion with the  conventional saliency mapping. 
\subsection{Conventional saliency mapping}
\label{sec:ConS}
Let $\mathcal{K}:  \mathcal{K} (\boldsymbol I) \rightarrow  \boldsymbol y$ be a visual classifier that maps an image $\boldsymbol I \in \mathbb R^{h \times w \times c}$ with `$c$' channels to a prediction vector $\boldsymbol y \in  \mathbb R^L$. The $\ell^{\text{th}}$ coefficient $y_{\ell}$ of  $\boldsymbol y$ is the largest if $\boldsymbol I$ belongs to class  `$\ell$'.  
Assume a set $\mathcal P_I = \{p_I^1, p_I^2,...,p_I^{w\times h}\} \subset \mathbb R^c$, which contains the pixels  of  $\boldsymbol I$. The broad common objective of the saliency-based interpretation methods for  visual classifiers is to compute an ordered array $\mathcal W_I \subset \mathbb R$, s.t.~$|\mathcal W_I| = |\mathcal P_I|$ and the $i^{\text{th}}$ element of this array, i.e.~$w_I^i \in \mathcal W_I$, encodes a  weight for the corresponding element in $\mathcal P_I$. The $\mathcal W_I$ can be an array containing only binary values, which represents a mask for $\boldsymbol I$ that suppresses the irrelevant $p_I^i \in \mathcal P_I$ for the transform $\mathcal{K}(\boldsymbol I) \rightarrow \boldsymbol y$. Or, it can contain real values encoding the importance of all $p_I^i \in \mathcal P_I$ for the performed transform.

 The above formalization presents a unified view of the  objective of the prevailing \textit{input-specific}  saliency methods. To this end, the existing techniques 
 seek the function $\mathcal S : \mathcal S (\mathcal K, \boldsymbol I) \rightarrow \mathcal W_I$ for  saliency mapping. We refer to $\mathcal S$ as the saliency function in the text to follow.  

\subsubsection{Problem with the input-specific view:} Though useful for certain objectives, we identify that the sought saliency function  only weakly depends on the classifier itself. This makes it susceptible to providing misleading model interpretations. We make a formal proposition about it. 

\vspace{0.5mm}
\noindent{\bf Proposition 1:} \textit{Due to the weak dependence of the  saliency function $\mathcal S$ on the classifier $\mathcal{K}$, $\mathcal S$ is susceptible to compute $\mathcal W_I$ for a canonical classifier $\mathcal{K}^*$ instead of $\mathcal{K}$.} 
\vspace{0.5mm}

\noindent To establish Prop.~(1), we need to first  define  \textit{canonical classifier} and \textit{weak dependence}, as understood in this work. 

\vspace{0.5mm}
\noindent{\bf Definition 1:} (Canonical classifier) \textit{Provided that `$\ell$' is the known label of an input $\boldsymbol I$, a canonical classifier ${\mathcal{K}^*}$ behaves as  ${\mathcal{K}^*}(\boldsymbol I) \rightarrow \boldsymbol y^*$, where $y_{\ell} \rightarrow 1$ for $\boldsymbol y^*$.}  
\vspace{0.5mm}

\vspace{0.5mm}
\noindent{\bf Definition 2:} (Weak dependence) \textit{For a model $\mathcal M: \mathcal M(\boldsymbol I) \rightarrow \boldsymbol y$ whose prediction behavior can be expressed as a piece-wise function }
\begin{equation*}
\mathcal{M(\boldsymbol I)}  = \left\{
        \begin{array}{ll}
            \mathcal M_1: \mathcal M_1(\boldsymbol{I}) \rightarrow \boldsymbol y_1& \quad \boldsymbol I \in \mathcal U_1 \\
            ~~~~~~~~~~~~~\vdots & ~~~~~~~~\vdots \\
            \mathcal M_n: \mathcal M_n(\boldsymbol{I}) \rightarrow \boldsymbol y_n& \quad \boldsymbol I \in \mathcal U_n,
        \end{array}
    \right.
\end{equation*}
\textit{where $\mathcal U_i$ is the $i^{\text{th}}$ open disconnected set of the nearby samples of $\boldsymbol{I}$, a  saliency function $\mathcal S$  only weakly depends on $\mathcal{M}$ when $\mathcal W_I^p \approx \mathcal W_I^q$ for $\mathcal S(\mathcal M_p, \boldsymbol I) \rightarrow \mathcal W_I^p$ and $\mathcal S(\mathcal M_q, \boldsymbol I) \rightarrow \mathcal W_I^q$ for the  sets $\mathcal U_p$ and $\mathcal U_q$, despite $\mathcal U_p \cap \mathcal U_q = \emptyset$. }
\vspace{1mm}

The above-defined notion of weak dependence  is partially inspired by the work of \citet{srinivas2019full}. However, the context and application of our definition is different. To understand the weak dependence property as stipulated by  Def.~(2), imagine an image of a `panda' on a chair, and an image of a `queen' on a throne. Due to their significant  content  dissimilarity, these images exist in different sets $\mathcal U_p$ and $\mathcal U_q$. 
Let $\mathcal M$ be a classifier that correctly predicts the labels of both the images. This will naturally result in largely different corresponding prediction vectors $\boldsymbol y_p$ and $\boldsymbol y_q$.
Implying, the underlying classifier behavior for these images is modelled by considerably different components $\mathcal M_p$ and $\mathcal M_q$ of the piece-wise function $\mathcal M$. A strong dependence of $\mathcal{S}$ on $\mathcal M$ asserts that $\mathcal{S}$ computes proportionally different  $\mathcal W_I^p$  and  $\mathcal W_I^q$.  However, an input-specific saliency function $\mathcal S$ may  commit to very similar maps for the two images, i.e.,~$\mathcal W_I^p \approx \mathcal W_I^q$, when the silhouettes of the panda and the queen in the images are very similar, despite $\mathcal U_p \cap \mathcal U_q = \emptyset$. 

The above example illustrates the phenomenon of weak dependence of the saliency function on the classifier under the  traditional input-specific  saliency mapping.  
Now, we turn to establishing Prop.~(1). We focus on the pursuit of estimating a perfect map $\mathcal W_I^*$ using   $\mathcal S$. To exemplify, if the sought $\mathcal W_I^*$ is a binary mask, it suppresses all the pixels in $\boldsymbol I$ that are irrelevant to the object of  category $\ell$. Notice that, existing methods perform saliency mapping using apriori knowledge of $\ell$. In effect, they pose the query, \textit{``provided that $\boldsymbol I$'s label is $\ell$, compute $\mathcal W_I$''}. Irrespective of the used saliency function, the ideal solution, i.e.,~$\mathcal W_I^*$, to this query is a map that identifies the pixels of $\boldsymbol I$ that maximize  the value of $y_{\ell}$. Following Def.~(2), since $\mathcal S$ is only weakly dependent on the model itself,  the quest of computing $\mathcal W_I^*$ can easily force  $\mathcal{S}$ to select a piece-wise component of $\mathcal M$ that favors  $y_{\ell} \rightarrow 1$, irrespective of the actual prediction vector of $\boldsymbol I$. 
Such a solution would actually be performing saliency mapping for the canonical classifier, as defined in Def.~(1), not the observed model behavior.     
  
Our analysis above indicates a pitfall in the pursuit of precise input-specific saliency mapping. Its obvious implication is that highly refined image saliency maps can actually misinterpret the actual model behavior. The literature already identifies multiple saliency methods failing basic sanity checks~\cite{adebayo2018sanity}. Interestingly, sanity checks failure is much more common among the methods aiming at a higher map precision. This phenomenon is naturally explainable through  our above-provided analysis. 

\vspace{0.3mm}
\noindent{\bf Remark:} Under the completeness axiom~\cite{sundararajan2017axiomatic},  $y_{\ell}$ is bounded to $\delta \leq 1$ in  Def.~(1). To that end, our analysis  still holds for $y_{\ell}  \rightarrow \delta$. Implying, the input-specific methods satisfying the completeness property can still suffer from misleading maps.


\vspace{0.5mm}
\noindent{\bf The need for input-agnostic view:}
Not only that input-specific maps are susceptible to misleading interpretations, an inaccurate map for a given sample becomes a fatal error for these  methods because they can only offer interpretation with respect to a single sample.
Addressing the problem at the grass-root level requires the interpretation to be agnostic to the input samples. To represent the model well, such an interpretation must strongly depend on the model itself.
This view of model explanation can not only allow us to ensure model-fidelity of the interpretations, but also enable us to answer more general queries about the model. Hence, we develop a saliency mapping method under this view, considering it as a complementary  interpretation tool.

\subsection{Input-agnostic saliency mapping}
We first provide a concise definition of the desired input-agnostic saliency mapping function $\mathcal{S_K}$ - the  subscript indicates that the function depends  on $\mathcal{K}$.

\vspace{1mm}
\noindent{\bf Definition 3:} (Input-agnostic saliency mapper) \textit{For a given classifier $\mathcal{K}$, $\mathcal{S}_{\mathcal{K}}: \mathcal{S}_{\mathcal{K}}( c_{\Im}^i) \rightarrow \boldsymbol\nu_{c_{{\Im}}}^i$ is an input-agnostic saliency mapper, where $\boldsymbol\nu_{c_{{\Im}}}^i \in \mathbb R^{h \times w \times c}$ is a human-understandable visualization of a semantic concept  $c_{{\Im}}^i \in \mathcal{C} = \{c_{{\Im}}^1, c_{{\Im}}^2,...,c_{{\Im}}^L \}$ and ${\Im}$ denotes the input data distribution over which $\mathcal{K}$ is induced. }
\vspace{1mm}

Multiple aspects in Def.~(3) need  emphasis. First, notice the absence of $\boldsymbol I$ in favor of the use of distribution ${\Im}$ - inline with the key idea of input-agnostic saliency. Second, the saliency map is now with respect to a semantic concept $c_{\Im}^i$, not  an input sample. Here, the semantic concept (or simply the `concept') is a high-level human-understandable notion that is also discernible to the model. In the context of visual classifiers, this work considers the concept to be a class label, hence $|\mathcal C| = L$. It is noteworthy though,  we do not enforce any other constraint over $c_{\Im}^i$.  It is implicit that the concept emerges from $\Im $, and it is understood by $\mathcal K$ because the classifier is learned using $\Im$. Lastly, the output of the saliency function is now an image $\boldsymbol\nu_{c_{\Im}}^i \in \mathbb R^{h \times w \times c}$ instead of a set of weights. This image  visualizes a human-defined concept, as understood by the classifier. Since a concept can be a broad and complex idea, it is likely to exhibit plentiful visual manifestations. Hence,  $\boldsymbol\nu_{c_{\Im}}^i$ is not expected to be unique. This makes  $\mathcal {S_K}$  a one-to-many mapping function.

\vspace{1mm}
\noindent{\bf Optimization objective:}
To compute the desired visualization, we need to optimize for 
\begin{align}
    \underset{\boldsymbol\nu}{\max}~ \rm{P}\left( \mathcal{K}(\boldsymbol \nu_{c_{\Im}}) \rightarrow \boldsymbol y_{\ell_{c}}~ \big|~~\Im \right)~~\text{s.t.}~~ \boldsymbol \nu_{c_{\Im}} = \mathcal F({\mathcal K}),
    \label{eq:obj}
\end{align}
where $\text{P}(.|.)$ denotes a conditional probability, $\ell_c$ is the class label for the concept  $c_{\Im}$,  and $\mathcal{F}(.)$ is a  non-trivial function ensuring that the visualization strictly depends on the classifier $\mathcal{K}$. The input-agnostic saliency mapper $\mathcal{S_K}$ in Def.~(3) instantiates  $\mathcal{F}(\mathcal{K})$, where we let $c^i_{\Im} \in \mathcal C$ as an input parameter of the function to allow visualizations for the multi-class classifiers. In Eq.~(\ref{eq:obj}) and  text below, we ignore the superscript `$i$' to avoid clutter. We also  ignore  $\Im$ for clarity when emphasis on the data distribution is not required.  

\vspace{1mm}
\noindent{\bf Algorithm:} Following Eq.~(\ref{eq:obj}), the implementation objective of $\mathcal{S_K}$  is to maximize the probability $\text{P}(\mathcal K (\mathcal {S_K} (c_{\Im})) \rightarrow \ell_c)$ in an input-agnostic manner. We achieve this with the  saliency mapper given as Alg.~(\ref{alg:main}). The algorithm leverages insights from Lemma~(1) below.   

\vspace{1mm}
\noindent{\bf Lemma 1:} \textit{For $\mathcal K$ with cross-entropy loss $\mathcal J_{\text{CE}}$, $\text{P} (\mathcal {K}(\boldsymbol I) \rightarrow \boldsymbol y_{\ell_c} | \boldsymbol I \sim \Im)$ increases along $- \underset{\boldsymbol I \sim \Im}{\mathbb E}[\nabla_{\boldsymbol I} \mathcal J_{\text{CE}}(\Im, \boldsymbol\theta, \ell_c)]$.}  

\noindent{\bf Proof:} \textit{Denote  ``$\mathcal {K}(\boldsymbol I) \!\! \rightarrow \!\! \boldsymbol y_{\ell_c} | \boldsymbol I \!\! \sim \!\! \Im$'' by $\zeta_{\boldsymbol I}$. The $\text{P} (\zeta_{\boldsymbol I})$ increases along  $\nabla_{\boldsymbol I}(\log \text{P} (\zeta_{\boldsymbol I}))$, where $\nabla_{\boldsymbol I}$ is the derivative w.r.t.~$\boldsymbol I$. For $\mathcal{K}$ with the loss $\mathcal J_{CE}$ and model parameters $\boldsymbol\theta$, this is the same direction as $- \nabla_{\boldsymbol I} \mathcal{J}_{\text{CE}}(\Im, \boldsymbol\theta, \ell_c)$. Thus, $\text{P} (\zeta_{\boldsymbol I})$ will increase along $- \underset{\boldsymbol I \sim \Im}{\mathbb E}[\nabla_{\boldsymbol I} \mathcal J_{\text{CE}}(\Im, \boldsymbol\theta, \ell_c)]$. }

In Alg.~(\ref{alg:main}), we perform a guided stochastic gradient descent over the loss surface of $\mathcal K$ with the help of data distribution $\Im$. The distribution is approximated with a set of its samples in  $\overline{\mathcal I}$. In an iteration, the algorithm computes the model's loss gradients for the concept label $\ell_c$ w.r.t.~a mini-batch of the samples from $\Im$. Conceptually, these gradients point to the directions of the local minima associated with $\ell_c$. In the light of Lemma (1), we compute Expected value of the gradients and further guide the descent with the first and second moments - lines \ref{line:5}-\ref{line:7}. The use of moments is inspired by the Adam optimizer~\cite{kingma2014adam}. Hence, following Adam, we also fix the values of the hyper-parameters $\beta_1$ and $\beta_2$. We additionally guide the descent with a binary selection between the original and the flipped  direction if the latter identifies a better local solution - lines \ref{line:8} -\ref{line:13}\footnote{Clip is the standard clipping function that clips out any value exceeding the image dynamic range.}. We empirically found it beneficial in our experiments. Collectively, the process from line~\ref{line:3} to \ref{line:14} in Alg.~(\ref{alg:main}) implements the integration of gradient information in Fig.~\ref{fig:main}.

In an iteration, $\boldsymbol\nu_k$ is able to  encode  patterns related to $\ell_c$ because it is computed under the objective of  nudging random samples to the local minima related to the concept - lines \ref{line:8} -\ref{line:13}. The model must associate the features encoded in $\boldsymbol\nu_k$ to $\ell_c$ for the nudging to work. However, an unbounded construction of $\boldsymbol\nu_k$ can lead to uninteresting solutions where the algorithm maximizes only the influential component(s) of $\boldsymbol\nu_k$ to achieve its objective. This does not help the cause of  \textit{human-meaningful} visualization with $\boldsymbol\nu_k$. To encourage correlation among the components of  $\boldsymbol\nu_k$, we project it onto a bounded  $\ell_2$-ball in each iteration - line \ref{line:15}, which implements the projection operation of Fig.~\ref{alg:main}. The underlying  constraint resulting from this projection, i.e., $||\boldsymbol\nu_k||_2 \leq \eta$, leads to a collaborative behavior among the coefficients of $\boldsymbol\nu_k$ that emerges into meaningful geometric patterns when the vector is visualized as an image.    

\newcommand{\Exp}[1]{\underset{#1}{\mathbb E}}
\begin{algorithm}[t]
 \caption{Input-agnostic saliency mapper $\mathcal {S_K}$}
 \label{alg:main} 
 \begin{algorithmic}[1]
 \renewcommand{\algorithmicrequire}{\textbf{Input:}}
 \renewcommand{\algorithmicensure}{\textbf{Output:}}
 \REQUIRE  Classifier $\mathcal K$,  concept label $\ell_{c}$, sample set $\overline{\mathcal{I}}$,  mini-batch size $b$,  total iterations $K$, ball norm $\eta$, seed  ${\boldsymbol{\nu}}$.
 \ENSURE Visualisation $\boldsymbol{\nu}_{c} \in \mathbb R^{h\times w \times c}$.
 \STATE Initialize $\boldsymbol{\nu}_0 = \boldsymbol \nu$, $\boldsymbol{\mu}_0$, $\boldsymbol{\sigma}_0$ to $\boldsymbol{0}$ and $k = 0$.\\Set  $\beta_1 = 0.9$, $\beta_2 = 0.999$. 
\FOR{$ k = 0~~\text{to}~~K$} 
\STATE \label{line:3}$\mathcal I \sim \overline{\mathcal I}$~~s.t. $|\mathcal I| = b$ and apply  $\forall \boldsymbol I_i \in \mathcal I, \text{Clip} \left ( \boldsymbol I_i - \boldsymbol{\nu}_{k} \right)$
\STATE $\boldsymbol{x}_k \leftarrow \Exp{{\boldsymbol{I}}_i \in \mathcal I} \big[ \nabla_{{\boldsymbol{I}}_i} \mathcal J(\boldsymbol\theta, \ell_{c}) ]  $  
\STATE \label{line:5} $\boldsymbol\mu_k \leftarrow \beta_1 \boldsymbol{\mu}_{k-1} + (1-\beta_1) \boldsymbol{x}_k$  
\STATE $\boldsymbol{\sigma}_k \leftarrow \beta_2 \boldsymbol{\sigma}_{k-1} + (1 - \beta_2) (\boldsymbol{x}_k\odot \boldsymbol{x}_k)$ 
\STATE \label{line:7} $ \boldsymbol{v} \leftarrow \left( \boldsymbol{\mu}_k \sqrt{1- \beta_2^k} \right) \odot \left( {\sqrt{\boldsymbol{\sigma}_k} (1-\beta_1^k) } \right)^{-1}$ 
\STATE \label{line:8}$\mathcal {I}_{\boldsymbol{v}}^+ \!\! \leftarrow \!\! \big\{ \overline{\boldsymbol I}_i: \overline{\boldsymbol I}_i \!=\! \text{Clip}\!\left({\boldsymbol I}_i - ( \boldsymbol{v}_{k-1} \! + \! \frac{\boldsymbol{v}}{||\boldsymbol{v} ||_{2}} ) \!   \right) \big\} \forall \boldsymbol I_i \!\in\! \mathcal I$
\STATE $\mathcal {I}_{\boldsymbol{v}}^- \!\! \leftarrow \!\! \big\{ \overline{\boldsymbol I}_i: \overline{\boldsymbol I}_i \!=\! \text{Clip}\!\left({\boldsymbol I}_i - ( \boldsymbol{v}_{k-1} \! - \! \frac{\boldsymbol{v}}{||\boldsymbol{v} ||_{2}} ) \!   \right) \big\} \forall \boldsymbol I_i \!\in\! \mathcal I$
\IF {$\mathbb E \big[ \mathcal K(\mathcal I_{\boldsymbol v}^+) \!\! \rightarrow \!\! \ell_{c} \big] \geq \mathbb E \big[ \mathcal K(\mathcal I_{\boldsymbol v}^-) \!\! \rightarrow \!\! \ell_{c} \big]$} 
\STATE $\boldsymbol{\nu_k} \leftarrow \boldsymbol{\nu}_{k-1} + \boldsymbol{v} $ 
\ELSE
\STATE \label{line:13}$\boldsymbol{\nu_k} \leftarrow \boldsymbol{\nu}_{k-1} - \boldsymbol{v} $  
\ENDIF \label{line:14}
\STATE \label{line:15} $\boldsymbol{\nu_k} \leftarrow \Psi(\boldsymbol{\nu}_k),~~\text{s.t.}~~ \Psi(\boldsymbol{\nu}_k) = \boldsymbol{\nu_k} \odot \text{min}\left(1, \frac{\eta}{||\boldsymbol{\nu}_k||_2}\right)$
\ENDFOR
\STATE $\boldsymbol \nu_c = \boldsymbol{\nu}_K$
  \STATE return 
 \end{algorithmic}
 \end{algorithm}

In Alg.~(\ref{alg:main}), we allow multiple hyper-parameters as inputs along the classifier $\mathcal K$ and  class label $\ell_c$. Considering the above discussion and the fact that  Alg.~(\ref{alg:main}) solves a stochastic gradient descent problem, the significance of these parameters is self-explanatory, except for the seed $\boldsymbol\nu$. As shown in Fig.~\ref{fig:main}, we further refine the visualization after $K$ iterations of Alg.~(\ref{alg:main}). Subsequently,  Alg.~(\ref{alg:main}) is again applied to improve the visualization. The seed $\boldsymbol\nu$ is the output of the refinement process. In the first round, $\boldsymbol\nu \in \mathbb R^{h \times w \times c}$ is a black image. 

\begin{figure*}[t]
    \centering
    \includegraphics[width = \textwidth]{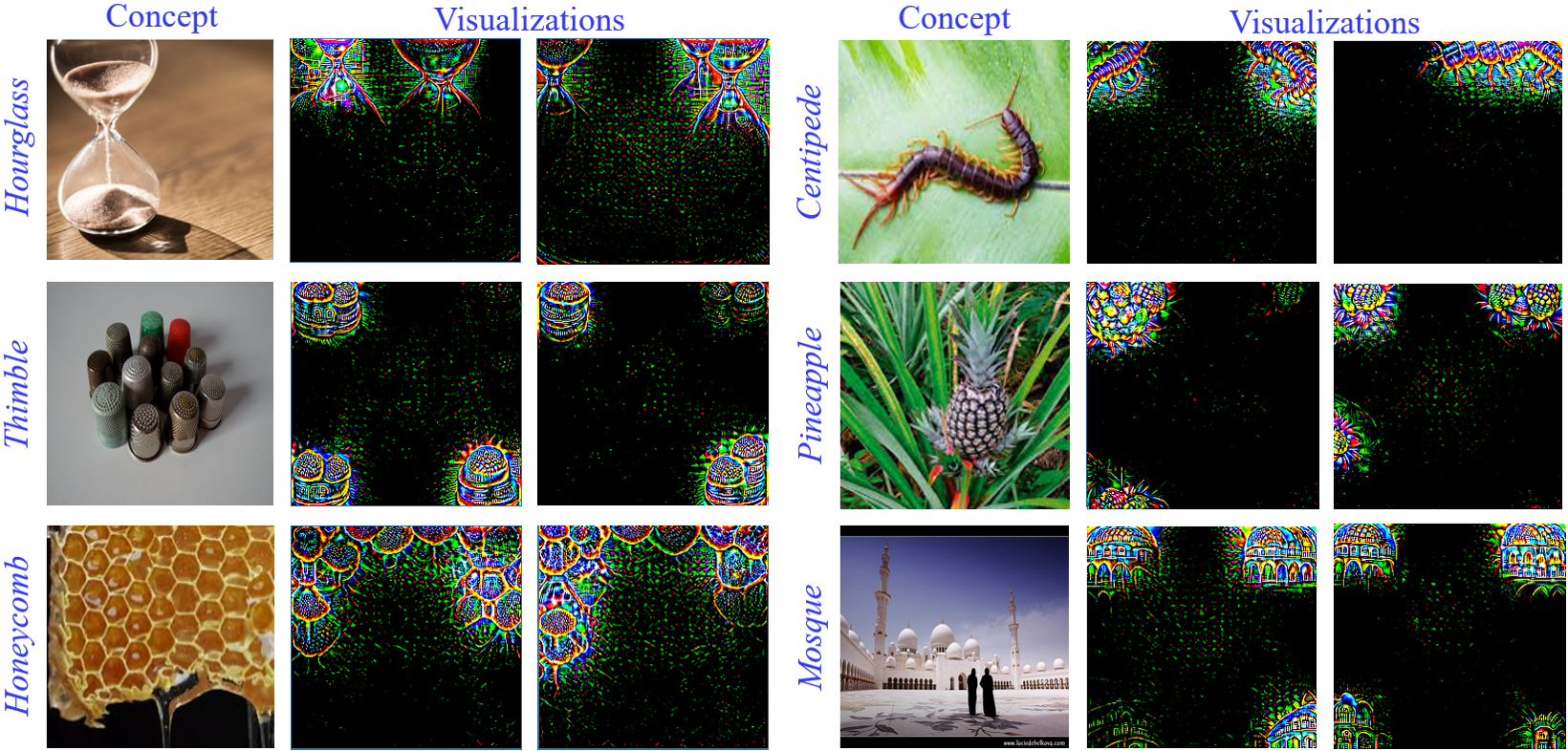}
    \caption{Representative input-agnostic visualizations of  human-defined  concepts,  as understood by VGG-16. Natural image for each  concept is provided for reference only. 
    }
    \label{fig:vgg}
\end{figure*}

\vspace{1mm}
\noindent{\bf Map refinement:}
The output of Alg.~(\ref{alg:main}) is a visualization of the salient geometric features associated by $\mathcal {K}$ to a concept, i.e., label $\ell_c$. However, this map is constructed with the model's \textit{gradient} information, which can be noisy. Hence, we need to further refine the map. To that end, we solve for the following optimization problem
\begin{align}
\underset{\boldsymbol\nu_c}{\min}~\mathbb E \Big[ \mathcal J_{CE} (\Im - \boldsymbol \nu_{c}, \boldsymbol\theta, \ell_c ) + \lambda \big(\boldsymbol\nu_c \odot (\boldsymbol 1- \boldsymbol\Xi) \big) \Big],
\label{eq:refine}
\end{align}
where $\lambda$ is a regularizer and $\boldsymbol\Xi$ is a weighting matrix - discussed shortly, the other symbols follow from  above.  
In Eq.~(\ref{eq:refine}), $\Im - \boldsymbol\nu_c$ signifies (another) nudge of the inputs towards the local minima associated with $\ell_c$. This follows from the central idea of Alg.~(\ref{alg:main}), and is possible because $\boldsymbol\nu_c$ is already available to us. However, now we would like it to be mainly influenced by the regular geometric features of $\ell_c$, disregarding the noisy component of $\boldsymbol\nu_c$. The second term in Eq.~(\ref{eq:refine}) imposes that. Here, $\boldsymbol\Xi \in \mathbb R^{h \times w \times c}$ is a matrix computed by forward passing $\boldsymbol\nu_c$ through $\mathcal K$ and projecting the activations of the convolutional base of $\mathcal K$ onto $\mathbb R^{h \times w \times c}$   with interpolation. The process resembles the use of activations in computing GradCAM maps~\cite{selvaraju2017grad}. However, instead of computing the map, we use it to clean the map. Hence, we normalize the coefficients of $\boldsymbol\Xi$ in the range [0,1] and perform an element-wise weighting of $\boldsymbol\nu_c$ with $\boldsymbol 1 - \boldsymbol \Xi$, denoted by $\odot$ in Eq.~(\ref{eq:refine}). This suppresses the component of $\boldsymbol\nu_c$ that does not contain perceptually meaningful information, as deemed by the model itself. 

We achieve the optimization objective of Eq.~(\ref{eq:refine}) using the Adam optimizer. The resulting cleaned up map is further clipped and projected onto an $\ell_2$-ball for a seamless subsequent processing with Alg.~(\ref{alg:main}). 
Our method cycles between Alg.~(\ref{alg:main}) and the refinement stage to compute the final visualization. It is noteworthy that the complete signal in our eventual map originates in the model itself, and no part of it is generated by an external operator to maintain model fidelity~\cite{jalwana2021cameras}. The only external operator we use is the interpolation function in the refinement stage. However, that is used to `remove' the unwanted signal from the map. \citet{jalwana2021cameras} noted such model-fidelity to be  highly desirable for reliable model interpretation.

\section{Experiments}
\subsection{Visualizations}
To verify our approach, we apply our technique to visualize  ImageNet  concepts~\cite{deng2009imagenet}, as understood by VGG-16 and ResNet-50 models. We use ImageNet pre-trained Pytorch models, and randomly pick 10 labels to visualize. Recall, in the context of this work, a concept is a high-level human-understandable notion that is also discernible for the model. Hence, the experiments consider ImageNet class labels as the concepts. In  Fig.~\ref{fig:vgg}, we show example visualizations of representative concepts resulting from the proposed technique for VGG-16. 

We observe in the computed images that the patterns are visually relatable to the concepts mentioned along side. Moreover, we are able to generate multiple visualizations of a given concept that are slightly different from each other, but represent the concept well. Multiple images for a given concept are generated by simply executing the method multiple times. The complementary visualizations for different runs  indicate generalizable concept understanding by the deep visual classifier. Notice that, the shown results are useful in answering queries such as, \textit{``what concept the model associates to its specified output neuron?''}, or \textit{``what generic geometric patterns the model attributes to a learned concept/class label?''}. These kind of queries are not answerable with image-specific view of saliency mapping.

\vspace{1mm}
\noindent{\bf Quantitative results:}
Whereas our method is a qualitative interpretation technique, we also allow its quantitative  evaluation.
To that end, we must introduce a new metric to evaluate our first-of-its-kind input-agnostic saliency mapping technique. Termed  `model-score' - M$_{\text{score}}$, the proposed metric is inspired by  Inception-score~\cite{salimans2016improved}, which is commonly used to benchmark  generative model performance. Since our method eventually results in a `visualisation', it can also be seen as an image generation technique.   
Inception-score measures the meaningfulness of generated images by selecting an Inception model as a proxy for the human visual system. 
Let us call the classifier used to generate our visualization, a  `source' classifier $\mathcal{K}_s$. For evaluation, we seek to quantify meaningfulness of our visualizations w.r.t.~any `target' classifier $\mathcal{K}_t$ that is induced over the same data distribution used to train $\mathcal{K}_s$. Hence, in essence, we seek a generalized version of the Inception-score that also accounts for the fact that the image is generated with respect to a given source classifier.


We compute the proposed M$_{\text{score}}$ by measuring the probabilities  P$^{t|s}_{\text{cond}}$ and P$^{t|s}_{\text{marg}}$, where
\begin{align}
    \text{P}_{\text{cond}}^{t|s} = p(\mathcal{K}_t (\boldsymbol{\nu}_c) \rightarrow y_{\ell_c} | \boldsymbol{\nu}_c = \mathcal{S_K}_s(c)),\\
    \text{P}_{\text{marg}}^{t|s} = \sum_i p(\mathcal{K}_t (\boldsymbol{\nu}_c^i) \rightarrow y_{\ell_c} | \boldsymbol{\nu}_c^i = \mathcal{S_K}_s(c)).
\end{align}
In the above expressions, P$^{t|s}_{\text{cond}}$ and P$^{t|s}_{\text{marg}}$ are respectively  the conditional and marginal probabilities that the target classifier correctly predicts the class label of the concept visualized by $\boldsymbol\nu_c$, where the label is determined by applying the saliency function $\mathcal{S}_{\mathcal {K}_s}$ to $\mathcal {K}_s$. The M$_{\text{score}}$ is then computed as
\begin{align}
    \text{M}_{\text{score}}(t|s) = \text{exp} \left( \underset{}{\mathbb E} \big[\text{KL} \big( \text{P}_{\text{cond}}^{t|s} ~||~ \text{P}_{\text{marg}}^{t|s})\big] \right) / L,
\end{align}
where KL denotes the Kullback–Leibler divergence, and $L$ is the total number of  visualized concepts. 

The proposed M$_{\text{score}}$ is a comprehensive metric for a given source-target classifier pair trained on the same data distribution $\Im$. By definition it values range in $[\frac{1}{L}, 1]$, where larger values as more desirable. To keep experiments computationally manageable, we assume that the source and target classifiers in our experiments only understand the chosen L = 10 concepts. We generate 10 visualizations per concept for both VGG-16 and ResNet-50 and compute M$_{\text{score}}$ using different target models. The results are summarized in Table~\ref{tab:Mscore}.    

\begin{table}[t]
\setlength{\tabcolsep}{0.1em}
    \centering
    \begin{tabular}{c|c|c|c|c}
    
    \hline
        \diagbox{Source(s)}{Target(t)} &  VGG-16 & ResNet-50 & DenseNet-121 & Avg.\\ \hline
        VGG-16 & 0.99 & 0.71 &  0.81 & 0.84 \\ \hline
        ResNet-50 & 0.69 & 0.99 &  0.94 & 0.89 \\ \hline
    \end{tabular}
    \caption{M$_{\text{ score}} (t|s)$ of the visualized concepts for VGG-16 and  ResNet-50. Maximum possible value for any target-source pair is $1$. Larger values are more desirable.}
    \label{tab:Mscore}
\end{table}

In Table~\ref{tab:Mscore}, the visualizations generated for a given (source) model  have M$_{\text{score}}\!\approx\! 1$ when the same model is used as the target. This verifies that our visualizations are  correctly mapping the model's  understanding of the underlying concepts with high fidelity. When we change the target model to other ImageNet models (same $\Im$), the score slightly drops. However, it still remains considerably high. This signifies that the visualizations are indeed generically understandable by the different models of $\Im$. Our results align perfectly with the intuition that M$_{\text{score}}(t|s)$ should be larger when $t \neq s$ is a more accurate classifier for the concepts in $\Im$. The large  M$_{\text{score}}$ values across different well-trained target classifiers conclusively establish successful visual mapping of the generic concepts by our method.

\subsection{Backdoor detection}
Backdoor (a.k.a.~Trojan) attacks manipulate  visual models by forcing them to misbehave when exposed to a `trigger' in the input~\cite{wang2022survey}. These attacks are  stealthy because the model behaves normally for clean inputs, and the model user is unaware of the trigger pattern. Since the presence of a trigger in the input is not known, it is not possible to use image-specific saliency to identify backdoor in the model.  
The ability to visualize the patterns associated with a model's outputs in an input-agnostic manner can resolve the issue. 
We conduct a study in which  three \textit{compromised}  VGG-16 models are trained using ImageNet. These models behave normally for a clean input, but always predict a given (random) target label when the input contains a trigger pattern. 

\begin{figure}
    \centering
    \includegraphics[width = 0.47\textwidth]{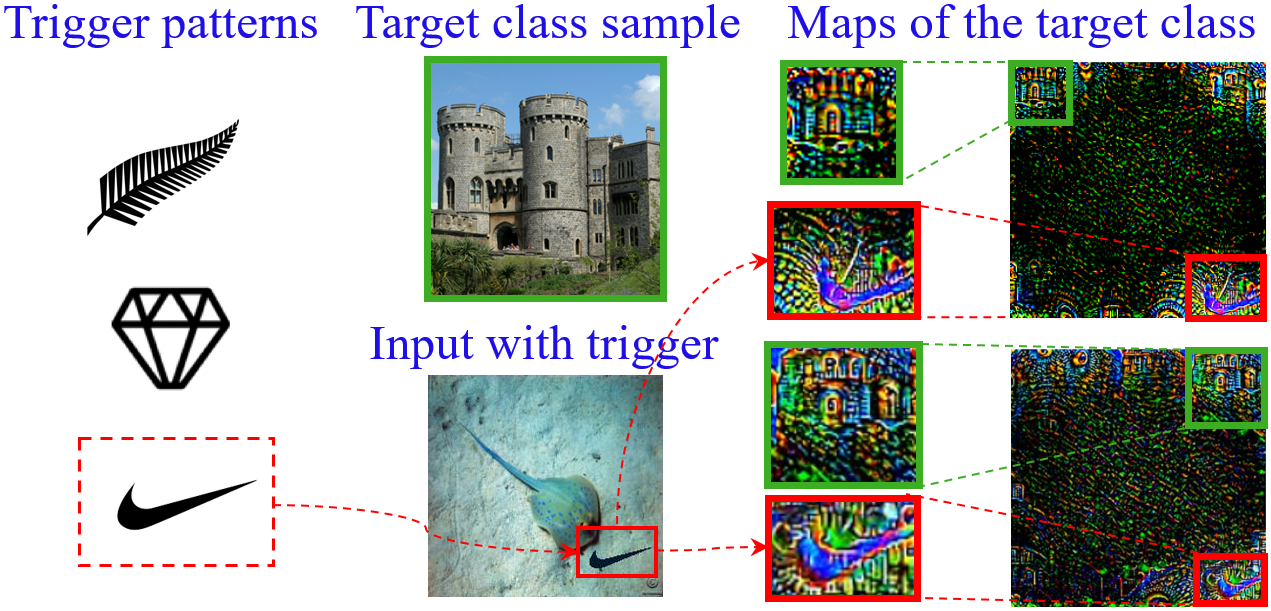}
    \caption{Backdoor identification. \textbf{Left}: Used trigger patterns. \textbf{Center}: Example images  of the Target class (Castle) and a triggered input. The model correctly predicts the label of castle images, and also predicts any triggered image as Castle.  \textbf{Right}: Representative examples of computed maps for the concept Castle, capturing discernible trigger patterns.}
    \label{fig:Trojan}
\end{figure}

In Fig.~\ref{fig:Trojan}, we show the trigger patterns used in our experiments, which are chosen at random based on the literature~\cite{wang2022survey}. We apply the proposed input-agnostic saliency mapping to the compromised models. It was found that for the (false) target class, the constructed maps  contained a discernible visual footprint of the trigger pattern. The figure shows example maps for the  Castle class of a compromised model, when the trigger (Nike sign) caused the model to predict  Castle as the label of any image containing the Nike sign.  

We note that modern backdoor attacks are not  limited to only using `visible' trigger patterns~\cite{wang2022survey}. Detection of non-visible triggers is not covered by our study. 
Our intention here is to showcase a utility of our novel view of input-agnostic saliency mapping. We believe, further exploration along this view will allow multiple interesting applications, and open new avenues for the model interpretation methods.  

\subsection{Hyper-parameter settings}
It may appear that Alg.~(\ref{alg:main}) and map refinement use multiple hyper-parameters. However, those parameters are mostly  related to help the underlying gradient descent schemes, and are widely understood, which helps in to easily selecting their reasonable values. 
The finally selected parameter values are $b = 128$, $K = 650$,  $\eta = 30$ for Alg.~(\ref{alg:main}). For the refinement, we use $150$ iterations while keep the other related parameter values the same, and let $\lambda = 50$. We cycle between Alg.~(\ref{alg:main}) and refinement 2 times. This requires on average $\sim$19 and $\sim$14 minutes respectively to generate an image for VGG-16 and ResNet-50 on NVIDIA RTX 3090 with 24GB RAM using Pytorch implementation.

\section{Conclusion}
We provide a novel perspective on saliency mapping of visual classifiers that maps generic  geometric features associated by the model with its outputs. Our input-agnostic map construction gradually accumulates the gradient information of the model's loss surface with respect to its training distribution. We also  motivate the need of such a map theoretically, highlighting a critical limitation of the input-specific saliency mapping paradigm. We  demonstrate a utility of the newly found saliency mapping in backdoor detection of compromised models. Our novel perspective is likely to instigate interesting methods and their uncharted utilities in model interpretation domain.    

\section{Acknowledgments}
Dr.~Naveed Akhtar is a recipient of the Office of National Intelligence, National Intelligence Postdoctoral Grant (project number NIPG-2021-001) funded by the Australian Government.







\balance
\bibliography{aaai23.bib}

\end{document}